\documentclass[final]{elsarticle}

\usepackage{lineno}
\usepackage{subfigure,gensymb,graphicx,booktabs}
\usepackage[abs]{overpic}
\usepackage{textcomp,latexsym,tikz,array}
\usepackage{booktabs,threeparttable}
\usepackage[colorlinks]{hyperref}
\usepackage[ruled,vlined,linesnumbered]{algorithm2e}
\usetikzlibrary{matrix}
\graphicspath{{figs/}}
\newcolumntype{L}[1]{>{\raggedright\arraybackslash}p{#1}}
\modulolinenumbers[1]

\journal{arXiv}

\biboptions{authoryear}

\begin{document}

\begin{frontmatter}

\title{Design mining microbial fuel cell cascades}

\author[csct]{Richard~J.~Preen\corref{correspondingauthor}}

\cortext[correspondingauthor]{Corresponding author}
\address[csct]{Department of Computer Science and Creative Technologies}
\address[bbc]{Bristol BioEnergy Centre, Bristol Robotics Laboratory \\ University of the West of England, Bristol, BS16 1QY, UK}
\ead{richard2.preen@uwe.ac.uk}

\author[bbc]{Jiseon~You}
\author[csct]{Larry~Bull}
\author[bbc]{Ioannis~A.~Ieropoulos}
 
\begin{abstract}
	Microbial fuel cells (MFCs) perform wastewater treatment and electricity production through the conversion of organic matter using microorganisms. For practical applications, it has been suggested that greater efficiency can be achieved by arranging multiple MFC units into physical stacks in a cascade with feedstock flowing sequentially between units. In this paper, we investigate the use of computational intelligence to physically explore and optimise (potentially) heterogeneous MFC designs in a cascade, i.e. without simulation. Conductive structures are 3-D printed and inserted into the anodic chamber of each MFC unit, augmenting a carbon fibre veil anode and affecting the hydrodynamics, including the feedstock volume and hydraulic retention time, as well as providing unique habitats for microbial colonisation. We show that it is possible to use design mining to identify new conductive inserts that increase both the cascade power output and power density.
\end{abstract}


\end{frontmatter}


\section{Introduction}

Microbial fuel cells (MFCs) are energy transducers that convert organic matter, including waste, directly into electricity using microorganisms. The available organic matter in MFCs varies, ranging from simple substances (e.g., acetate, glucose, and butyrate) to complex substances (e.g., municipal wastewater, brewery wastewater, and urine)~\citep{Pant:2010,Zhang:2011,Ieropoulos:2012}. In addition to the advantages of direct energy conversion, MFCs are environmentally friendly, recovering energy from waste.

Scale-up through the physical stacking and electrical connection of multiple MFC units is essential for practical applications of the technology since the energy density of an individual MFC unit is relatively low~\citep{Aelterman:2006,Ieropoulos:2008}. For treating wastewater, stacking MFC units in a cascade manner has been considered as a potential replacement of the trickling filter treatment process~\citep{Winfield:2012}. This type of process is very efficient for small to medium sized volumes of wastewater treatment where large areas of land are unavailable. Furthermore, it does not require high power to run since aeration or mixing is unnecessary. However, when connecting multiple MFC units electrically, issues such as cell reversal can be problematic~\citep{Oh:2007,Ieropoulos:2010a}. Several factors, which include substrate depletion, immaturity of anode biofilms, and inappropriate external electrical load, could lead to cell reversal~\citep{Oh:2007,Ieropoulos:2010a,Harnisch:2009,An:2014}. Among these, feedstock distribution within cascades is particularly important since feedstock composition changes throughout the stack, which in essence changes the internal resistance.

Design mining is the use of computational intelligence techniques to iteratively search and model the attribute space of physical objects evaluated directly through rapid prototyping to meet given objectives. It enables the exploitation of novel materials and processes without formal models or complex simulation, whilst harnessing the creativity of both computational and human design methods. A sample-model-search-sample loop creates an agile/flexible approach, i.e., primarily test-driven, enabling a continuing process of prototype design consideration and criteria refinement by both producers and users. We have recently demonstrated the success of the approach to discover new wind turbine designs~\citep{Preen:2014,Preen:2015,Preen:2016}. In this paper, we explore the application of design mining for MFC cascade design.

A recent review of the evolution of physical systems can be found in \cite{Preen:2015} and is therefore not covered here. The remainder of this paper is organised as follows. We begin with a brief overview of MFC cascades and a tuneable model that can be used to explore aspects of coevolution. Method and results are then presented for simulations of MFC cascade coevolution. Subsequently, we present results from a physical experiment used to optimise designs of conductive structures that are 3-D printed and inserted into each MFC unit in a cascade, augmenting a carbon fibre veil anode and affecting the hydrodynamics, including the feedstock volume and hydraulic retention time, as well as providing unique habitats for microbial colonisation.

\section{Background}

\subsection{Cascades of microbial fuel cells}

MFCs convert the chemical energy of feedstock into electricity through the metabolic activity of microorganisms. They usually consist of two compartments, an anode and a cathode, separated by an ion-permeable material. In the anode, microorganisms oxidise organic matter (fuel) and release CO$_2$, electrons, and protons. Electrons produced in the anode, flow to the cathode via an external circuit as the result of electrophilic attraction from the cathode electrode, whilst protons migrate from the anode to the cathode through the separator between the two compartments. The electrons and protons subsequently combine with oxygen (final electron acceptor) and this reduction reaction completes the circuit~\citep{Li:2008,Chae:2008}. The quantity of electrons flowing through the external circuit is the electricity being produced, i.e., current.

For MFC scale-up in terms of power generation, two distinct approaches have been suggested. The first is to increase the size of an individual MFC~\citep{Logan:2010}. The second is to build a multitude of relatively small MFCs connected electrically~\citep{Ledezma:2013,Ieropoulos:2016}. In the case of the second approach, the output of an MFC system is amplified by the number of MFC units employed, similar to how batteries can be connected together. Scale-up is also vital to treat certain volumes of wastewater. It is generally accepted that continuous flow is more favourable for both stable power generation and the treatment of large volumes of wastewater.

In a cascade system, the original feedstock is supplied to the first MFC unit positioned on the top of the cascade. Its effluent flows to the unit immediately below; the effluent of the first MFC therefore becomes the substrate for the downstream MFC. This is the same for third, fourth, etc. The performance of each MFC, in terms of the amount of substrate utilised, reproduction of anodic microorganisms or by-products from substrate utilisation, can therefore have a significant effect on other MFCs, despite not being connected hydraulically. For example, downstream MFCs are likely to be fed with a lower concentration of feedstock compared with the MFCs upstream since some of the readily available substrates are depleted before arrival. When single or multiple units experience substrate depletion, the power output is consequently reduced. In the long term, this can affect the anodic biofilm, causing the units to permanently underperform.

Hence, providing sufficient quality feedstock to each MFC unit in a stack is essential. One solution is to supply high concentrations of feedstock to cascades, to ensure all of the units have sufficient fuel. However, for maximum substrate utilisation and therefore maximum waste treatment efficiency, this may not be the best approach. Other operating parameters such as the flow rate, which consequently determines HRT, can be tuned to achieve both efficient energy production and waste treatment. Given that a cascade has only one source of feedstock flow, different HRT for individual MFC units in the cascade can be formed by changing the volume of each unit. \cite{Walter:2016} have recently shown that the power density of a cascade can be increased by reducing the size of downstream MFC units.

Many studies have investigated different design parameters, including the distance between the anode and cathode, and the surface area of both electrodes~\citep{Liu:2006,Oh:2006,Scott:2015}. However, studies of MFC reactor design optimisation embracing these parameters together are rare. This is particularly true for MFC cascade systems. In this respect, additive manufacturing technology, also known as 3-D printing, can be a very useful tool to test novel MFC architectures as well as electrodes and separators relatively quickly and easily~\citep{Ieropoulos:2010b,Calignano:2015,You:2017}. The design optimisation of individual MFC units will improve the individual performance, which results in performance enhancements to the whole system in terms of waste treatment efficiency and level of power output. This can also save the cost of materials, building and system footprint.

\subsection{Parallelism}

Genetic algorithms~\citep{Holland:1975} are population-based search and optimisation techniques that are inherently parallel. For example, multiple offspring can be evaluated simultaneously. The multi-population approach of coevolutionary algorithms~\citep{Husbands:1991} presents even more ways in which parallelism can be used. For example, combinations of offspring from multiple species populations can be evaluated simultaneously. Consequently, there is a growing body of work using graphics processing units to parallel process evolutionary algorithms, thereby reducing the wall-clock time required to find suitable solutions~\citep{Liu:2015}. Analogous to the use of graphics processing units for parallel hardware processing, in this paper we use multiple 3-D printers and multiple physical testing equipment to perform parallel evaluation of new MFC designs.

\subsection{Adaptive population sizing}

Most evolutionary algorithms fix the number of parents and offspring throughout optimisation. However, adapting the number of offspring with respect to evolutionary convergence has long been suggested as a means to improve the progress per fitness function evaluation~\citep{Hansen:1995}. The underlying concept is to adjust the population size to the smallest necessary to maintain sufficient search (genetic) diversity, thereby conserving the number of fitness function evaluations per generation. Recently, \cite{LaPort:2015} have shown that adapting the parental population size with respect to mutation and fitness rates can optimise the search diversity and thus improve fitness and convergence. See~\cite{Karafotias:2015} for an overview of parameter control in evolutionary algorithms, including the population size.

\subsection{The NKCS model}

\cite{Kauffman:1991} introduced the abstract NKCS model to enable the study of various aspects of coevolution. In their model, an individual is represented by a genome of $N$ (binary) genes, each of which depends epistatically upon $K$ other randomly chosen genes in its genome. Thus increasing $K$, with respect to $N$, increases the epistatic linkage, increasing the ruggedness of the fitness landscapes by increasing the number of fitness peaks, which increases the steepness of the sides of fitness peaks and decreases their typical heights. Each gene is also said to depend upon $C$ randomly chosen traits in each of the other $X$ species with which it interacts, where there are $S$ number of species in total. The adaptive moves by one species may deform the fitness landscape(s) of its partner(s). Altering $C$, with respect to $N$, changes how dramatically adaptive moves by each species deform the landscape(s) of its partner(s).

The model assumes all inter- and intragenome interactions are so complex that it is appropriate to assign random values to their effects on fitness. Therefore, for each of the possible $K+(X \times C)$ interactions, a table of $2^{K+(X \times C)+1}$ fitnesses is created for each gene, with all entries in the range [0,1], such that there is one fitness for each combination of traits. The fitness contribution of each gene is found from its table; these fitnesses are then summed and normalised by $N$ to give the selective fitness of the total genome for that species. Such tables are created for each species. 

See example in Fig.~\ref{fig:nkcs}; the reader is referred to \cite{Kauffman:1993} for full details. This tuneable model has previously been used to explore coevolutionary optimisation, for example in the comparison of partnering strategies~\citep{Bull:1997b}. We similarly use it here to systematically compare various techniques for the design mining approach.

\begin{figure}[t]
	\centering
	\small
	$N=3$~~~~$K=1$~~~~$C=1$~~~~$S=2$~~~~$X=1$\\
	\begin{tikzpicture}
		\node(a) at (0,0)[shape=circle,draw] {$n1$};
		\node(b) at (2,0)[shape=circle,draw] {$n2$};
		\node(c) at (1,-1)[shape=circle,draw] {$n3$};
		\node(astate) at (0,0.6)[] {1};
		\node(bstate) at (2,0.6)[] {0};
		\node(cstate) at (1,-0.4)[] {1};
		\node(label1) at (1,1.5)[] {Species $s1$};
		\node(d) at (5,0)[shape=circle,draw] {$n1$};
		\node(e) at (7,0)[shape=circle,draw] {$n2$};
		\node(f) at (6,-1)[shape=circle,draw] {$n3$};
		\node(dstate) at (5,0.6)[] {1};
		\node(estate) at (7,0.6)[] {1};
		\node(fstate) at (6,-0.4)[] {0};
		\node(label2) at (6,1.5)[] {Species $s2$};
		\draw [black, line width=1pt, ->, >=stealth]
		(a) edge node [above] {} (b)
		(c) edge node [above] {} (a)
		(b) edge node [above] {} (c)
		(d) edge node [above] {} (e)
		(e) edge node [above] {} (d)
		(e) edge node [above] {} (f);
		\draw [black, dashed, line width=1pt, ->, >=stealth]
		(d) edge [bend right=30] node [above] {} (a)
		(f) edge [bend right=5] node [above] {} (b)
		(f) edge [bend right=5] node [above] {} (c)
		(a) edge [bend left=40] node [above] {} (e)
		(b) edge [bend left=5] node [above] {} (d)
		(c) edge [bend right=10] node [above] {} (f);
		\node(em) at (0,-1.6)[] {};
	\end{tikzpicture}
\\
\setlength\tabcolsep{2pt}
	\begin{tabular}{ccc|c}
		\small
		\\
		\multicolumn{4}{c}{Species $s1$ gene $n1$} \\
		\hline
		$s1n1$ & $s1n3$ & $s2n1$ & fitness \\
		\hline
		0 & 0 & 0 & 0.57 \\
		0 & 0 & 1 & 0.12 \\
		0 & 1 & 0 & 0.09 \\
		0 & 1 & 1 & 0.16 \\
		1 & 0 & 0 & 0.44 \\
		1 & 0 & 1 & 0.66 \\
		1 & 1 & 0 & 0.33 \\
		\bf{1} & \bf{1} & \bf{1} & \bf{0.44}\\ [0ex]
	\end{tabular} 
	\hspace{5mm}
	\begin{tabular}{ccc|c}
		\small
		\\
		\multicolumn{4}{c}{Species $s1$ gene $n2$} \\
		\hline
		$s1n2$ & $s1n1$ & $s2n3$ & fitness \\
		\hline
		0 & 0 & 0 & 0.11 \\
		0 & 0 & 1 & 0.32 \\
		\bf{0} & \bf{1} & \bf{0} & \bf{0.68} \\
		0 & 1 & 1 & 0.30 \\
		1 & 0 & 0 & 0.19 \\
		1 & 0 & 1 & 0.77 \\
		1 & 1 & 0 & 0.21 \\
		1 & 1 & 1 & 0.23 \\ [0ex]
	\end{tabular}
	\hspace{5mm}
	\begin{tabular}{ccc|c}
		\small
		\\
		\multicolumn{4}{c}{Species $s1$ gene $n3$} \\
		\hline
		$s1n3$ & $s1n2$ & $s2n3$ & fitness \\
		\hline
		0 & 0 & 0 & 0.75\\
		0 & 0 & 1 & 0.42\\
		0 & 1 & 0 & 0.25\\
		0 & 1 & 1 & 0.28\\
		\bf{1} & \bf{0} & \bf{0} & \bf{0.13}\\
		1 & 0 & 1 & 0.58\\
		1 & 1 & 0 & 0.66\\
		1 & 1 & 1 & 0.91\\ [0ex]
	\end{tabular}
	\caption{The NKCS model: Each gene is connected to $K$ randomly chosen local genes (solid lines) and to $C$ randomly chosen genes in each of the $X$ other species (dashed lines). A random fitness is assigned to each possible set of combinations of genes. The fitness of each gene is summed and normalised by $N$ to give the fitness of the genome. An example NKCS model is shown above and example fitness tables are provided for species $s1$, where the $s1$ genome fitness is 0.416 when $s1=[101]$ and $s2=[110]$.}
	\label{fig:nkcs}
\end{figure}

\section{Simulations of MFC cascade coevolution}

\subsection{Methodology}

In this paper, 5 cascades composed of 4 MFCs are evaluated in parallel, where each cascade is physically duplicated and the average output is then used as fitness, i.e., there are 10 physical cascades. The fittest cascade is re-run in the subsequent test cycle to ensure that new designs are compared consistently. Thus, initially 5 different cascades are simultaneously evaluated to generate initial data, each composed of $S=4$ randomly created individuals. Thereafter, 4 new cascade evaluations occur in parallel on each test cycle. 

Initially we use the NKCS model to simulate MFC cascade coevolution and compare different evaluation strategies. To simulate the cascade flow it is assumed that the first species is epistatically dependant only on the second species (i.e., $X=1$), whereas the second and third species are linked to both neighbouring species (i.e., $X=2$) and the fourth species linked only to the third (i.e., $X=1$). An individual MFC is represented by $N=20$ genes. 

The new individuals produced for each test cycle are generated as follows. In the standard (1+4) algorithm, each species is evaluated sequentially by creating 4 offspring from the single fittest individual in the species and partnering each with the fittest individuals in the other species; see Algorithm~\ref{alg:1plus4}. In (1+1)$\times S$, each of the $S=4$ species produces a single offspring from the fittest member in parallel and each are partnered with the fittest individuals in the other species; see Algorithm~\ref{alg:1plus1x4}. In (1+4)-off, each species produces 4 offspring from the fittest member of that species in parallel and are partnered together for evaluation, i.e., $4 \times S$ offspring are created and tested in parallel; see Algorithm~\ref{alg:1plus4off}.

\begin{algorithm}[t]
    \small
	\SetAlgoLined%
	create and evaluate initial random designs\;
	\While{evaluation budget not exhausted}{
		\For{each species}{
			create 4 offspring using genetic operators\;
			partner with the fittest member in each other species\;
			evaluate the 4 cascades\;
			update the fittest design in each species 
		}
	}
	\caption{Coevolutionary algorithm $(1+4)$}
	\label{alg:1plus4}
\end{algorithm}
 
\begin{algorithm}[t]
    \small
	\SetAlgoLined%
	create and evaluate initial random designs\;
	\While{evaluation budget not exhausted}{
		\For{each species}{
			create 1 offspring using genetic operators\;
		}
		partner with the fittest member in each other species\;
		evaluate the 4 cascades\;
		update the fittest design in each species
	}
	\caption{Coevolutionary algorithm $(1+1)\times 4$}
	\label{alg:1plus1x4}
\end{algorithm}
 
\begin{algorithm}[t]
    \small
	\SetAlgoLined%
	create and evaluate initial random designs\;
	\While{evaluation budget not exhausted}{
		\For{each species}{
			create 4 offspring using genetic operators\;
		}
		partner with offspring in each other species\;
		evaluate the 4 cascades\;
		update the fittest design in each species
	}
	\caption{Coevolutionary algorithm $(1+4)$-off}
	\label{alg:1plus4off}
\end{algorithm}
 
Given the small initial population size, the same evaluation strategies are compared with versions where the population size of each species increases to a maximum of 50. That is, each algorithm runs as before, however tournament selection is used for parental selection (and replacement after the population reaches the maximum.) The tournament sizes are set to 80\% of the current population size. In all algorithms, the per allele mutation probability is 5\%, with a crossover probability of 0\%. All results presented are an average of 100 experiments consisting of 10 coevolutionary runs on 10 randomly generated NKCS functions.

\subsection{Results}

\begin{figure}[t]
	\centering 
	\includegraphics[width=\linewidth]{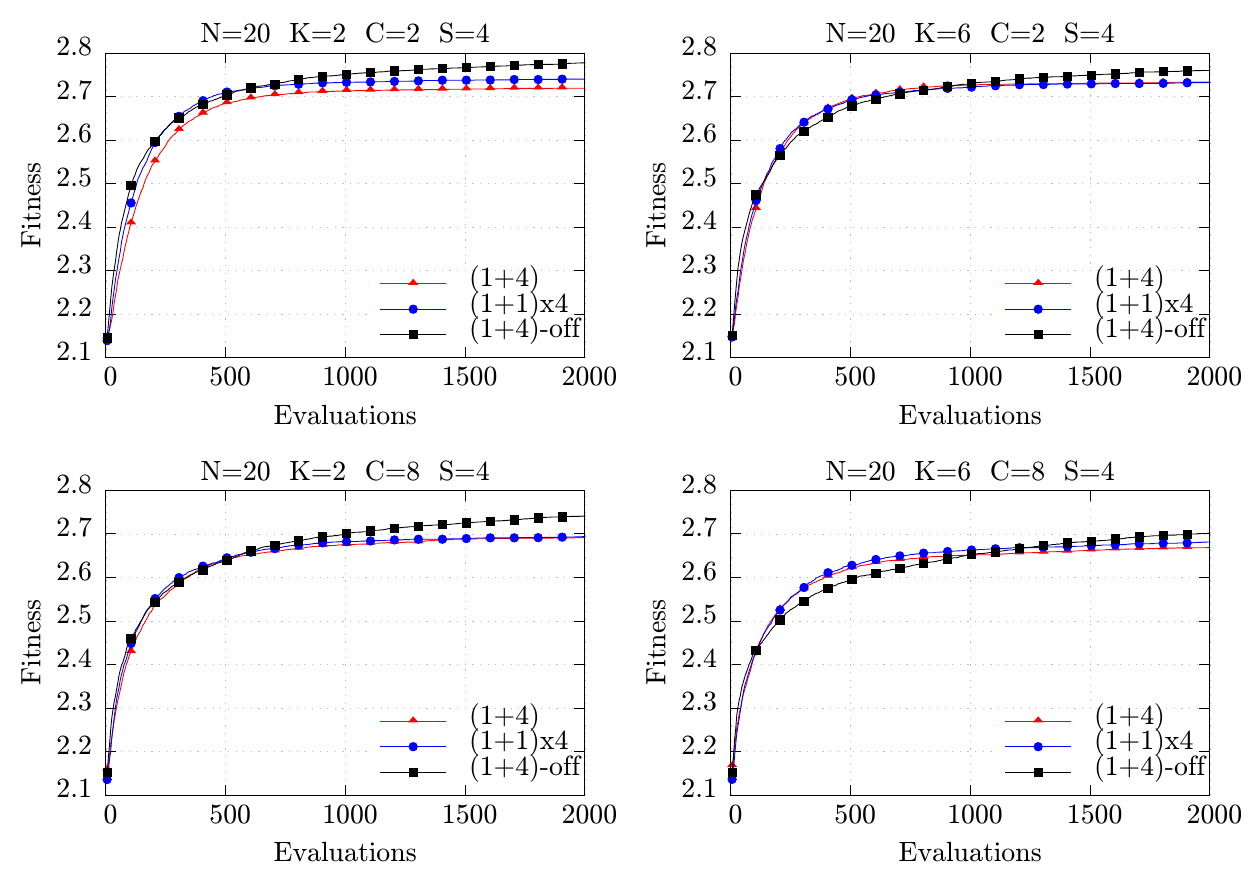}
	\caption{Simulations of cascade coevolution. Results are an average of 100 experiments consisting of 10 coevolutionary runs of 10 random NKCS functions.}
	\label{fig:share}
\end{figure}

\begin{figure}[t]
	\centering 
	\includegraphics[width=\linewidth]{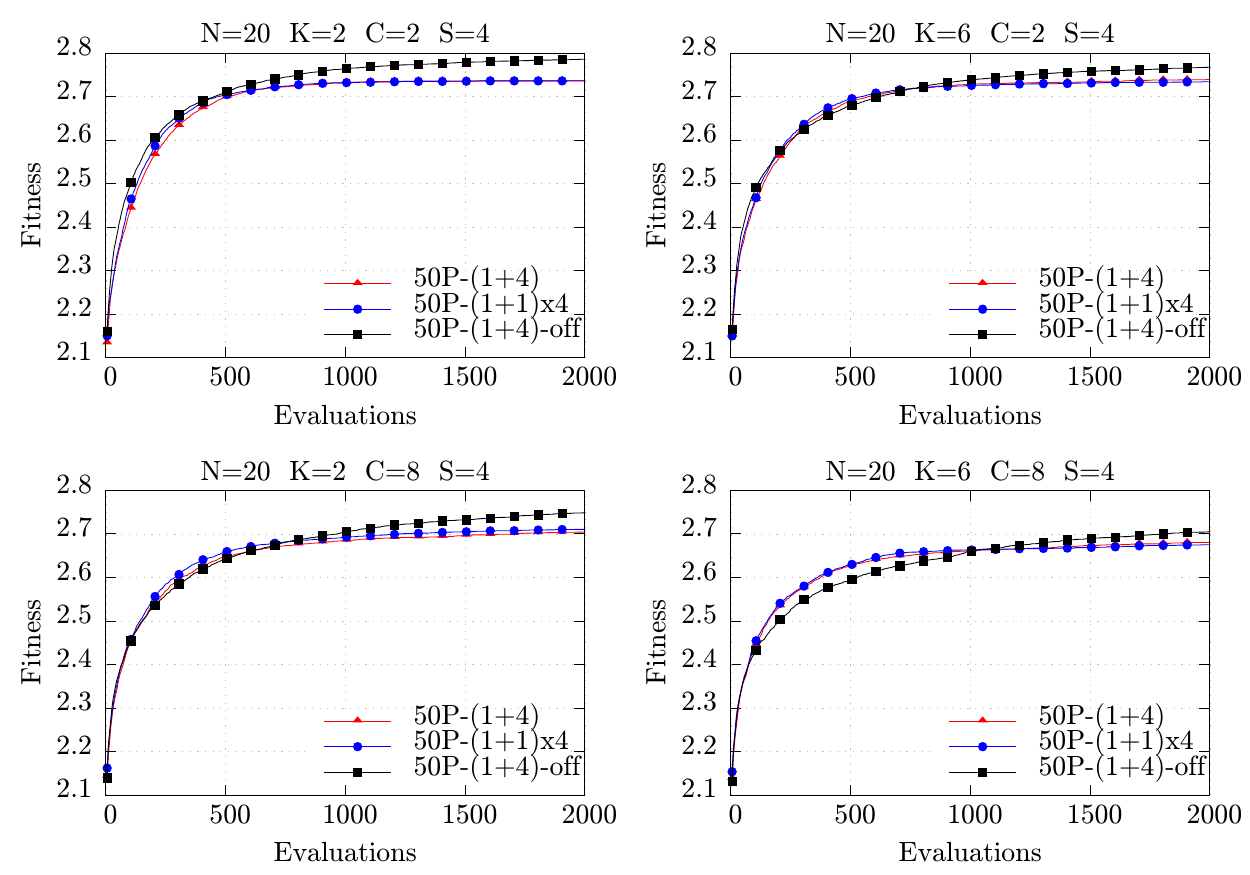}
	\caption{Simulations of cascade coevolution with expanding population sizes. Results are an average of 100 experiments consisting of 10 coevolutionary runs of 10 random NKCS functions.}
	\label{fig:share_50p}
\end{figure}

The performance of the different evaluation strategies are shown for 4 different $K$ and $C$ values in Fig.~\ref{fig:share}, each representing a different point in the range of inter- and intra-population dependence. The performance of the same strategies with an expanding population size are shown in Fig.~\ref{fig:share_50p}.

The results show that during the early stages of evolution there is little difference between the algorithms. For example, Table~\ref{table:sim_stats} shows that there is no significant difference after 400 evaluations when $K$ and $C$ are greater than 2. However, after 2000 evaluations, evaluating all offspring together is significantly greater for all values of $K$ and $C$, showing that this approach is able to search a wider design space while following increases in the fitness gradient. Expanding the population size appears to provide no performance increase to the algorithms; see Table~\ref{table:sim_ep_stats}, which shows that there are no significant differences.

\begin{table}[t]
	\caption{Simulation best fitnesses after 400 and 2000 evaluations (averages of 100). The mean is highlighted in boldface where it is significantly different from the (1+4) algorithm using a Mann-Whitney $U$ test at the 95\% confidence interval}
    \centering
    \begin{tabular}{l r r r }
		\toprule
		& $(1+4)$ & $(1+1)\times4$ & $(1+4)$-off \\
		\hline
		\multicolumn{4}{l}{After 400 evaluations:} \\
		$K2C2$ & 2.6593 & {\bf 2.6890} & {\bf 2.6805} \\
		$K2C8$ & 2.6224 & 2.6245 & 2.6163 \\
		$K6C2$ & 2.6714 & 2.6706 & 2.6510 \\
		$K6C8$ & 2.6012 & 2.6091 & 2.5735 \\
		\multicolumn{4}{l}{After 2000 evaluations:} \\
		$K2C2$ & 2.7190 & 2.7399 & {\bf 2.7772} \\
		$K2C8$ & 2.6916 & 2.6938 & {\bf 2.7407} \\
		$K6C2$ & 2.7331 & 2.7318 & {\bf 2.7598} \\
		$K6C8$ & 2.6684 & 2.6812 & {\bf 2.7013} \\
		\bottomrule
    \end{tabular}
    \label{table:sim_stats}
\end{table}
 
\begin{table}[t]
	\caption{Simulation best fitnesses with expanding population sizes after 400 and 2000 evaluations (averages of 100). The mean is highlighted in boldface where it is significantly different from the fixed population size version using a Mann-Whitney $U$ test at the 95\% confidence interval}
    \centering
    \begin{tabular}{l r r r }
		\toprule
		& 50P-(1+4) & 50P-(1+1)$\times$4 & 50P-(1+4)-off \\
		\hline
		\multicolumn{4}{l}{After 400 evaluations:} \\
		$K2C2$ & 2.6717 & 2.6840 & 2.6890 \\
		$K2C8$ & 2.6264 & 2.6385 & 2.6173 \\
		$K6C2$ & 2.6642 & 2.6724 & 2.6558 \\
		$K6C8$ & 2.6084 & 2.6097 & 2.5762 \\
		\multicolumn{4}{l}{After 2000 evaluations:} \\
		$K2C2$ & 2.7359 & 2.7361 & 2.7857 \\
		$K2C8$ & 2.7042 & 2.7102 & 2.7482 \\
		$K6C2$ & 2.7391 & 2.7336 & 2.7672 \\
		$K6C8$ & 2.6802 & 2.6746 & 2.7051 \\
		\bottomrule
    \end{tabular}
    \label{table:sim_ep_stats}
\end{table}
 
Since the physical testing of MFCs is a time consuming process, there is only a small evaluation budget available and therefore there appears to be no benefit from the additional exploration by evaluating all offspring simultaneously. Consequently, in the rest of this paper the coevolutionary algorithm $(1+1)\times4$ is used as this fits more naturally with the parallel evaluation performed with the physical test equipment and, although not statistically significant, the mean fitness values were greater for small numbers of evaluations.
 
\section{Physical MFC cascade coevolution}

\subsection{Methodology}

\begin{figure}[t]
	\centering 
	\begin{overpic}[width=12cm]{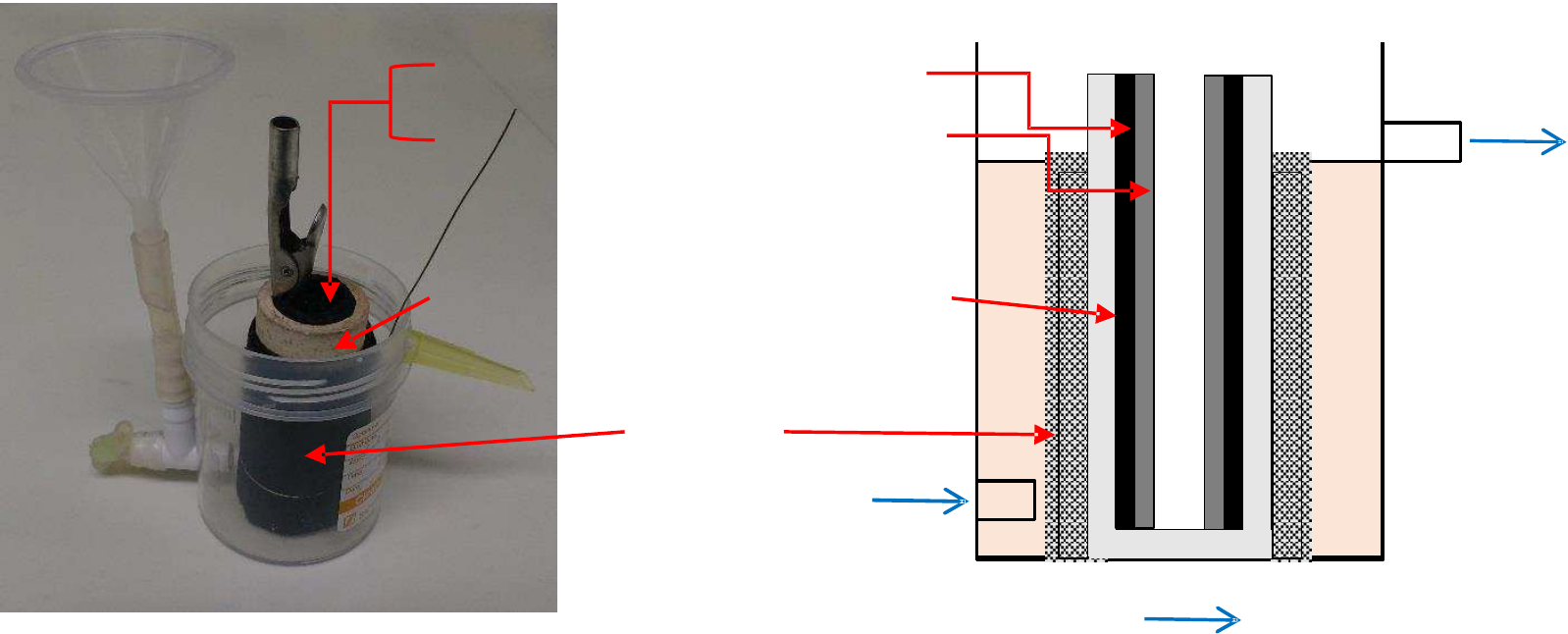}
		\put (100,120) {\scriptsize Activated carbon (cathode)}
		\put (100,105) {\scriptsize Gas diffusion layer (cathode)}
		\put (100,73) {\scriptsize Ceramic cylinder (separator)}
		\put (140,42) {\scriptsize Anode}
		\put (190,1) {\scriptsize Feedstock flow}
	 \end{overpic}
	\caption{MFC unit assembly.}
	\label{fig:mfc}
\end{figure}

\begin{figure}[t]
	\centering 
	\includegraphics[height=5.25cm]{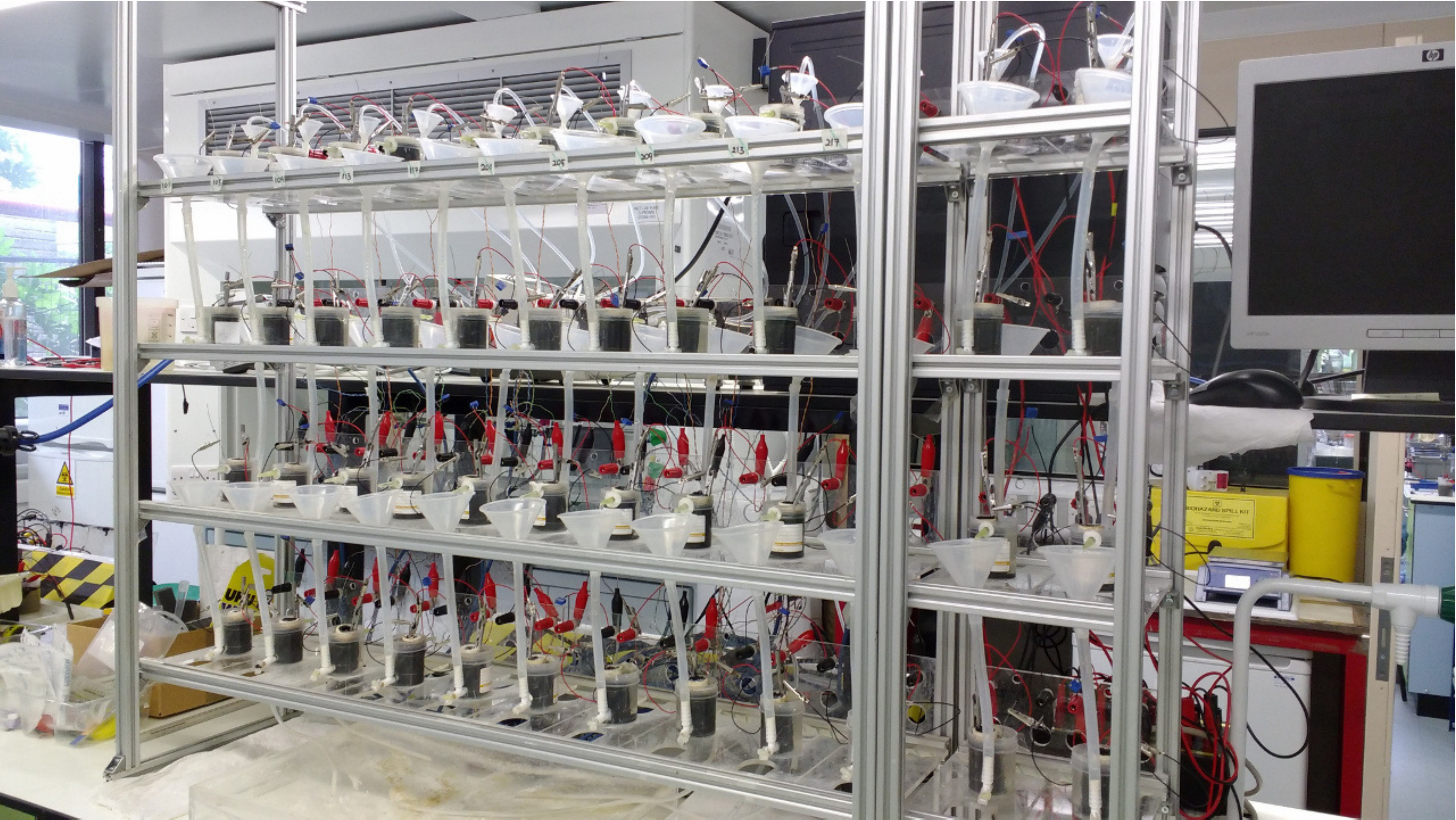}
	\begin{overpic}[height=5.25cm]{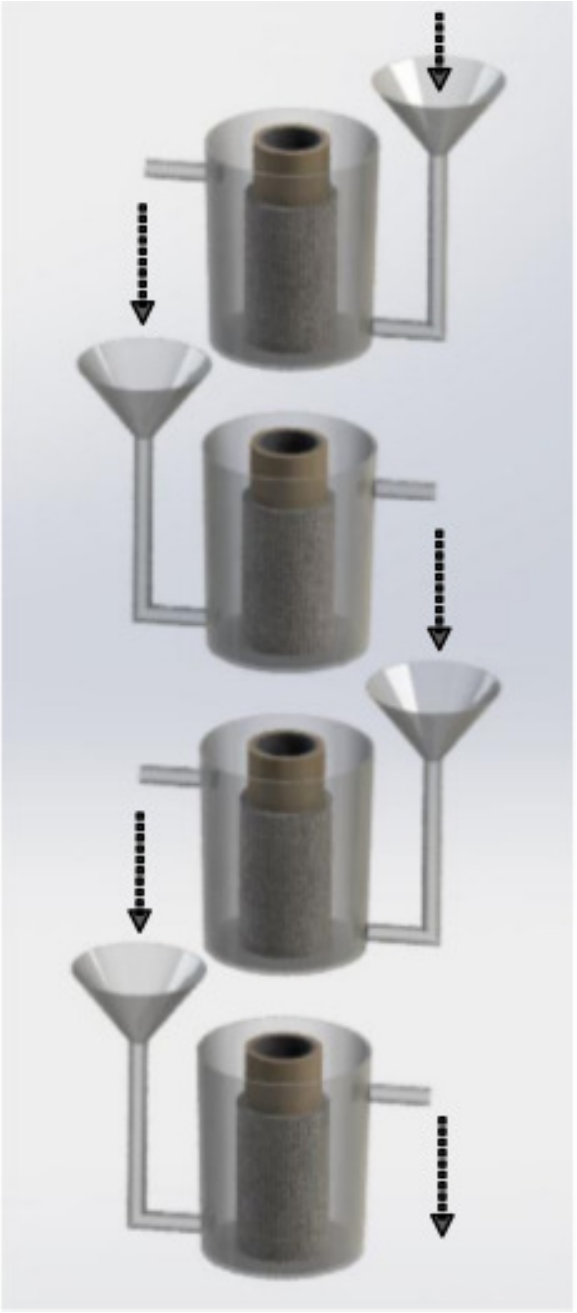}
		 \put (28,110) {\huge 1}
		 \put (28,76) {\huge 2}
		 \put (28,43) {\huge 3}
		 \put (28,7) {\huge 4}
	 \end{overpic}
	\caption{Experimental setup. 12 cascades of 4 MFC units (left) and cascade schematic (right). Numbers show each MFC unit position within a stack and arrows indicate direction of feedstock flow. 5 unique cascade designs and a control are evaluated in parallel, each duplicated for averaging.}
	\label{fig:setup}
\end{figure}

For this study, 12 cascades in a pair (1 pair as a reference) were run and each cascade consisted of 4 MFC units. For each MFC unit, a ceramic cylinder (inner diameter: 17~mm, thickness: 3~mm, height: 50~mm; EM80P; Anderman Industrial Ceramics Ltd., UK) was placed in a cylindrical plastic container, acting as the electrode separator. A carbon fibre veil (carbon loading: 20~g/m$^2$; Plastic Reinforcement Fabrics Ltd., UK) anode with size of 270~cm$^2$ (width: 30~cm, length: 9~cm) was wrapped around the ceramic separator and a hot-pressed activated carbon cathode electrode with a total surface area of 20~cm$^2$ (width: 4~cm, length: 5~cm) was placed inside the separator. Without a 3-D printed insert, this container held 44~mL volume of anolyte. Fig.~\ref{fig:mfc} shows one MFC reactor assembled with an outer anodic chamber and inner air-cathode compartment configuration. Each MFC unit was placed 15~cm apart in cascade manner. The experimental setup and a cascade schematic can be seen in Fig.~\ref{fig:setup}.

Cylindrical anode inserts were fabricated with a 3-D printer (Replicator~2; MakerBot Industries LLC, USA) using a conductive polylactic acid (PLA) based filament (Proto-pasta; ProtoPlant Inc., USA) at 0.3~mm resolution. The conductive filament is a compound of PLA, a dispersant, and conductive carbon black. The measured volume resistivity of 3-D printed conductive parts perpendicular to layers was 44~$\Omega\cdot$cm. See~\cite{You:2017} for further information on commercially available 3-D printed filaments for MFC anodes and membranes. This conductive PLA based filament was chosen to provide conductive inner structure as well as to create different volumes, inner shapes, and hydrodynamics in each MFC unit's anodic chamber. 

\begin{figure}[t]
	\centering 
	\includegraphics[width=3cm]{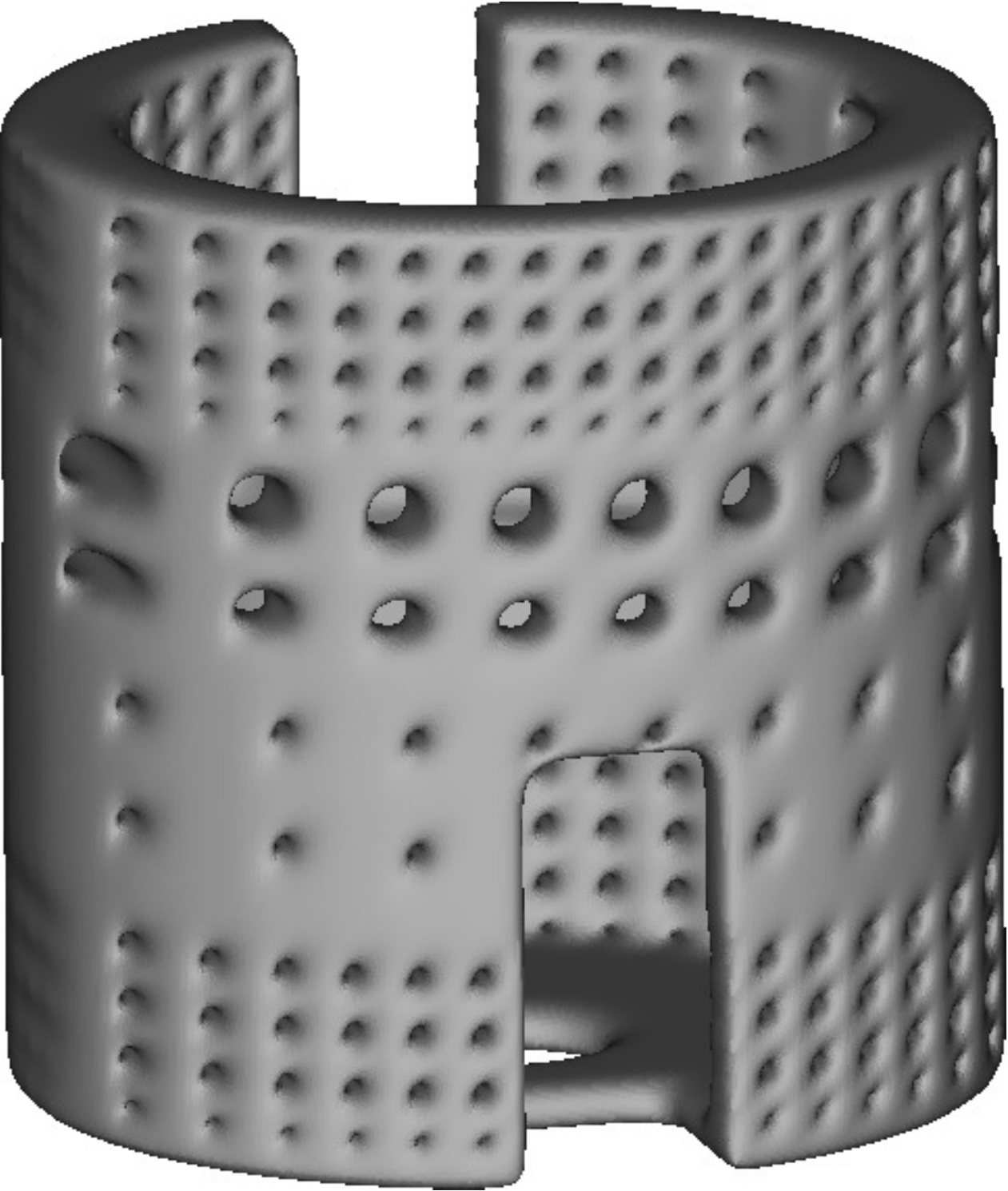} \hspace{1cm}
	\includegraphics[width=3cm]{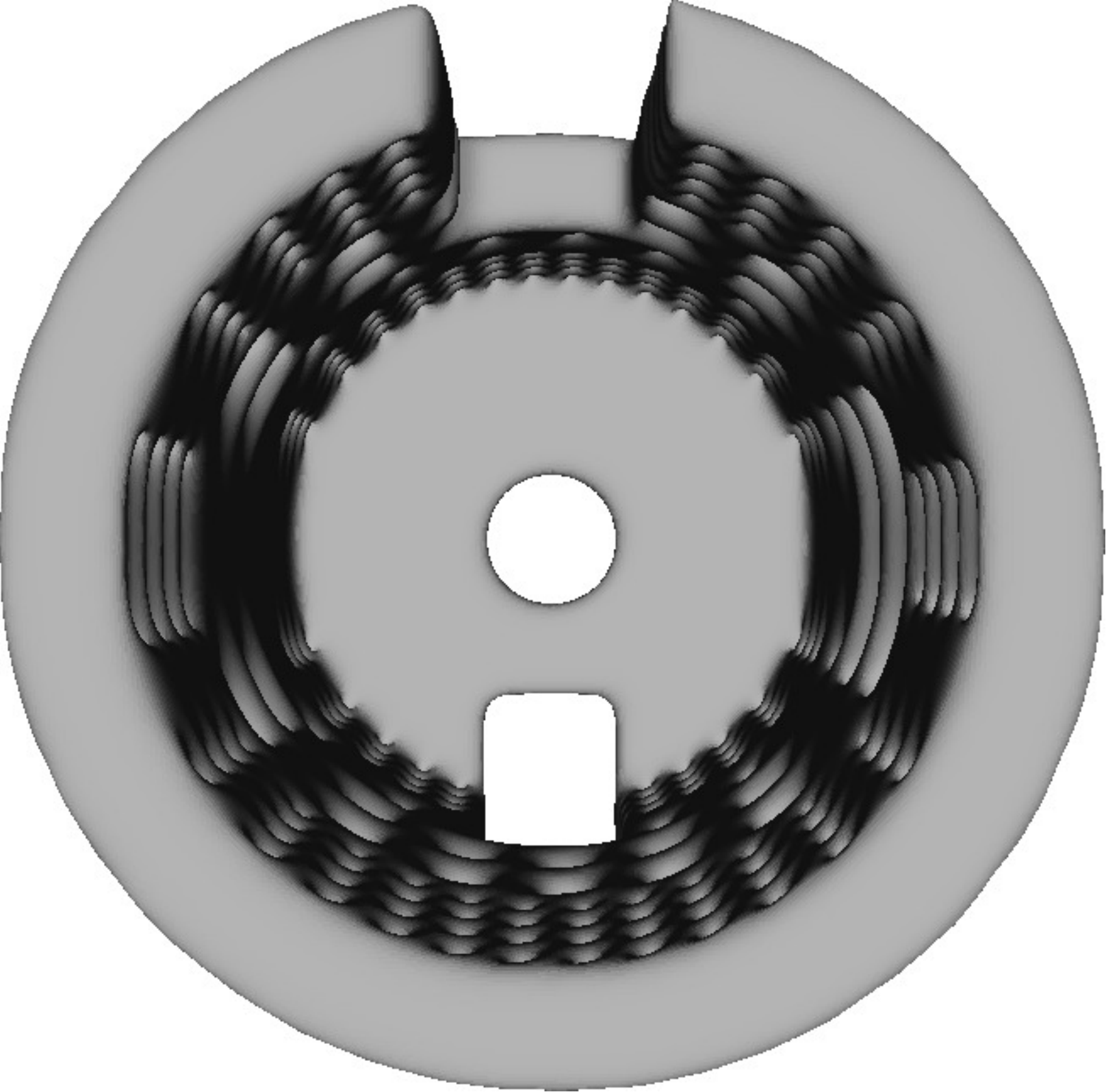}
	\caption{Example anodic chamber insert with 4 sections each of 10~mm height. Each section may vary the inner radius, diameter of holes, and hole spacing. Genome: [1,1,0,1,1,2,0,2,1,0,1,0]. Each insert has a 2~mm thick base with a 6~mm diameter hole in the bottom so that it lies flat in the centre of the MFC container, and includes rectangular spaces of 8~mm width for inlet (bottom half) and outlet (opposite side, top half). 50~min fabrication time.}
	\label{fig:insert}
\end{figure}

The total height of each insert was fixed at 40~mm in order to fit inside the MFC container. Each insert was divided into 4 sections of 10~mm height. Each section was encoded by 3 genes that represent the inner radius, diameter of holes, and distance between holes. Therefore, there are 12 genes in total. The outer radius of each section was fixed to 19.5~mm, i.e., the inner radius of the container. The minimum inner radius of each section was 14.5~mm and the maximum 17.5~mm. This was necessary to enable sufficient space for the ceramic separator, cathode, and carbon veil anode, and to ensure that a minimum sufficient amount of structural material was deposited for subsequent $z$-layers. Circular holes were created along the $y$-axis. Hole diameters were permitted from 0 to 3~mm, and distance between holes from 1 to 4~mm. Thus, each insert design is encoded as 12 integers (genes) in the range [0,3], each allele representing 1~mm increments between the permissible ranges. For example, if the first 3 genes are [0,1,1], the bottom section has an inner radius of 14.5~mm (14.5+0); hole diameters of 1~mm (0+1); and hole spacing of 2~mm (1+1). There are approximately 20,000 possible morphologies in each cascade position; see example in Fig.~\ref{fig:insert}. Offspring are created by copying and mutating 2 genes from the fittest individual in each species/position. A mutation event randomly increments or decrements the gene by 1 (mm).

The MFCs were inoculated with sewage sludge (Wessex Water Services Ltd., UK) and fed with synthetic wastewater~\citep{Winfield:2012} with 10~mM of acetate as a sole carbon source. Feedstock solution was provided continuously at 14~mL/hr to the first MFC units of the cascades, and effluent of the first units overflowed to the ones below. The MFCs were hydraulically disconnected. Initially 2~K$\Omega$ external resistors were connected to all MFCs, then the value of resistance changed to 1~K$\Omega$ for the first and second MFC units of a cascade. Power output of the MFCs was monitored continuously in volts (V) using a multi-channel Agilent 34972A DAQ unit (Agilent Technologies Inc., USA) every 5~min. For chemical oxygen demand (COD) analysis, effluent of 2~mL volume was taken from each MFC unit and filter-sterilised with 0.45~{\textmu}m syringe filters (Millex; Millipore UK Ltd., UK) prior to analysis. COD was determined using the potassium dichromate oxidation method (COD~MR test vials; Camlab Ltd., UK) and analysed with a photometer (Lovibond MD~200; The Tintometer Ltd., UK). The experiment was carried out in a temperature controlled environment at $22\pm2$ \degree C.

At the beginning of each week, all MFCs were run without 3-D printed inserts for 2 days. Power outputs during this time were used to normalise the stack outputs once inserts were added. 3-D printed inserts generated using Algorithm~\ref{alg:1plus1x4} were subsequently added into the MFC anodic compartments in order to create different inner volumes and structures. MFCs were run for an additional 2 days with inserts and the average power output of the cascade units during this period were used as fitness values. In the event of a tie, the design producing the highest power density became the winner. The experiment was run for 10 generations, with initial inserts generated randomly.

\subsection{Results}

\begin{figure}[t]
	\centering 
	\includegraphics[width=10cm]{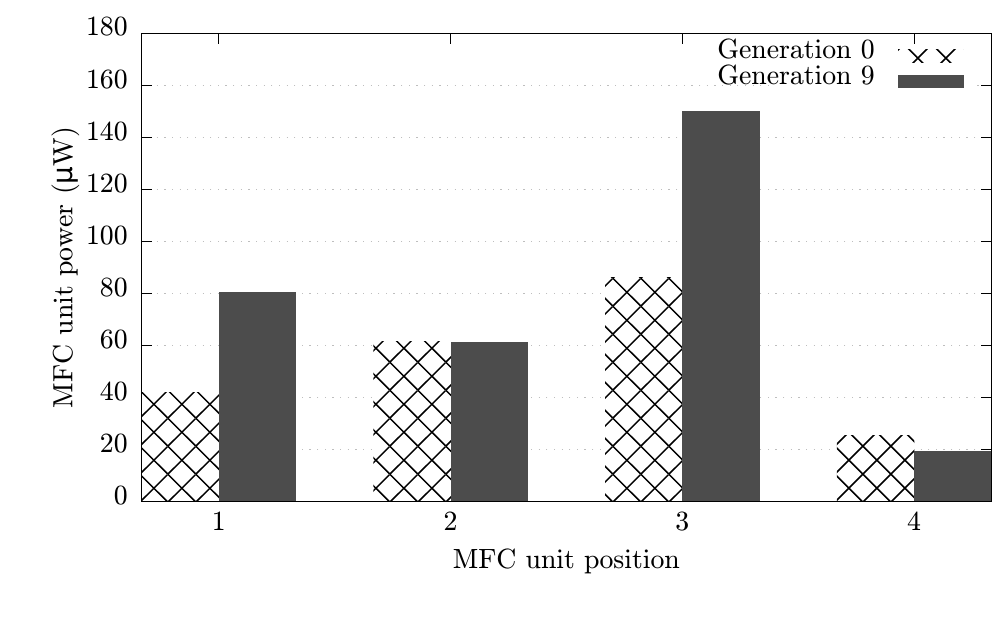}
	\caption{Power produced by each MFC unit in generation 0 (pattern) vs. generation 9 (solid).}
	\label{fig:power}
\end{figure}

\begin{figure}[t]
	\centering 
	\subfigure[Generation 0.] {
		\includegraphics[width=2.5cm]{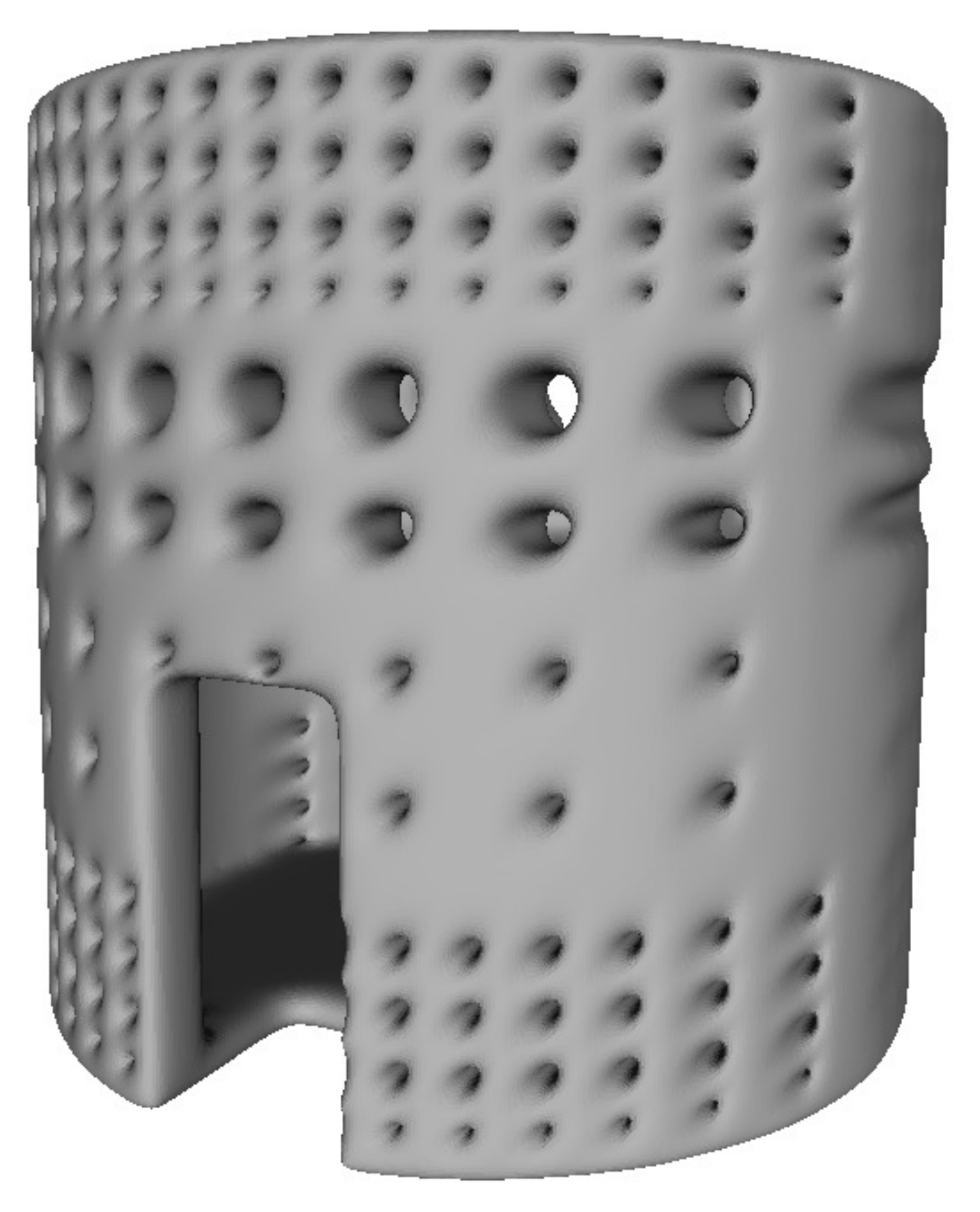}
		\includegraphics[width=2.5cm]{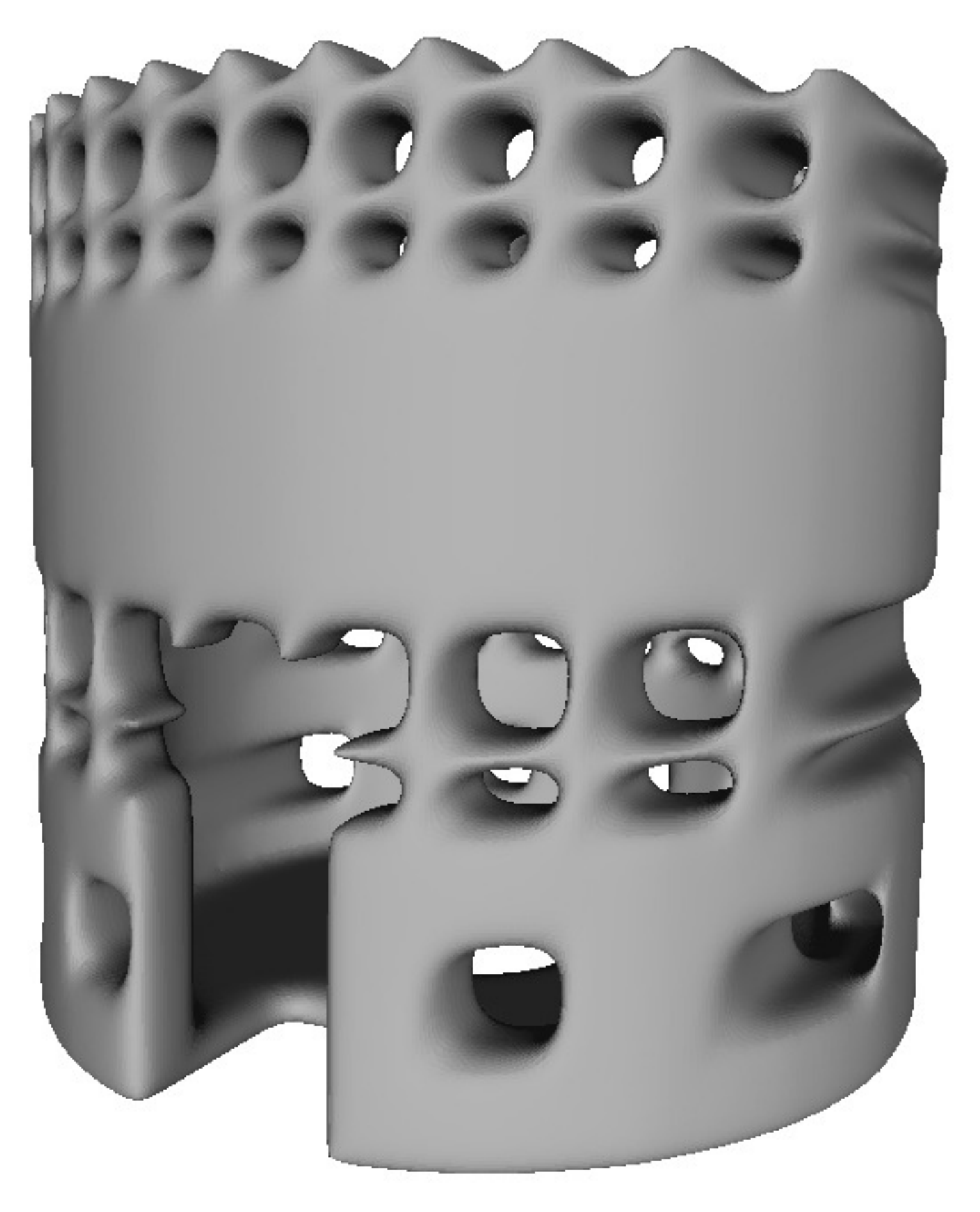}
		\includegraphics[width=2.5cm]{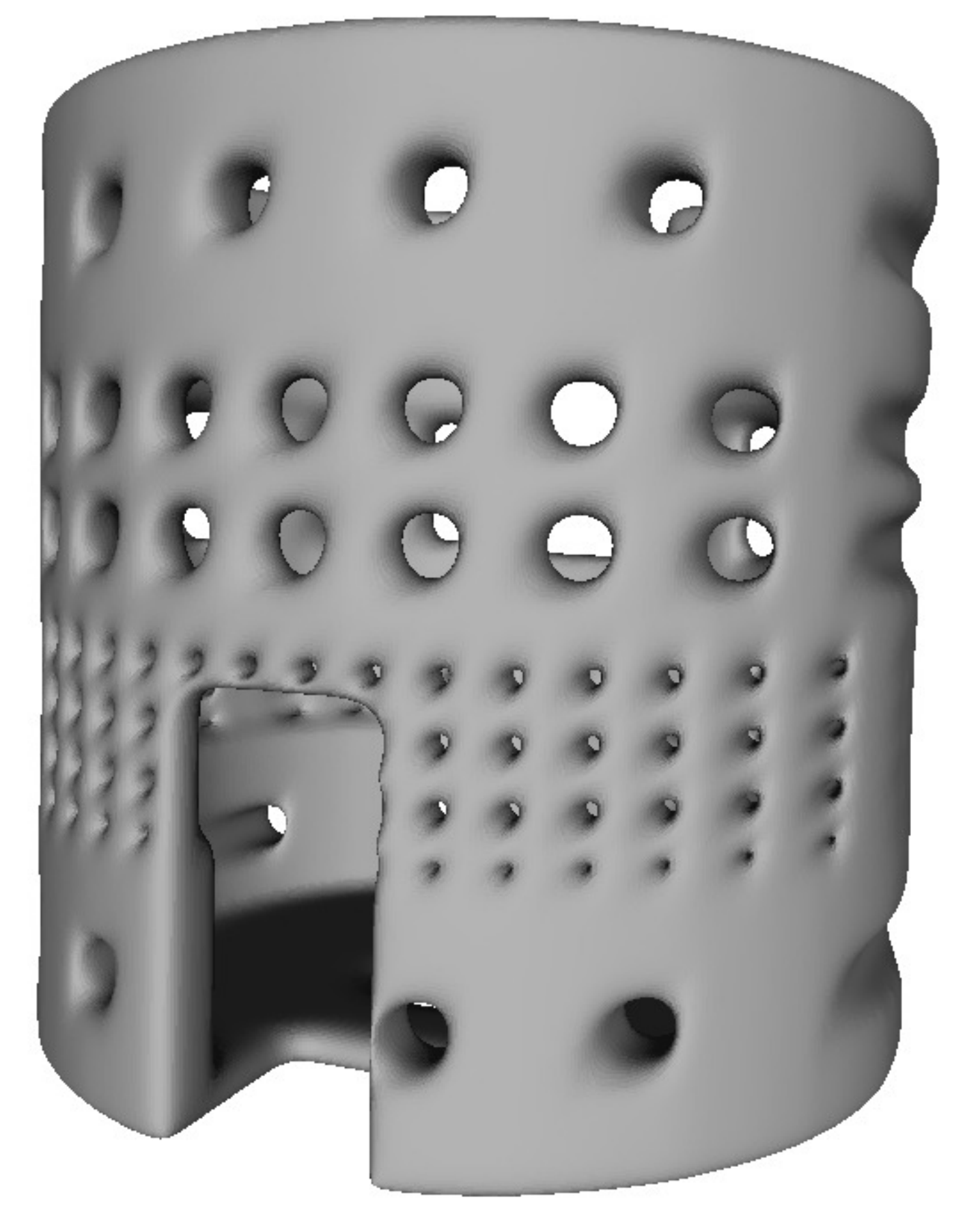}
		\includegraphics[width=2.5cm]{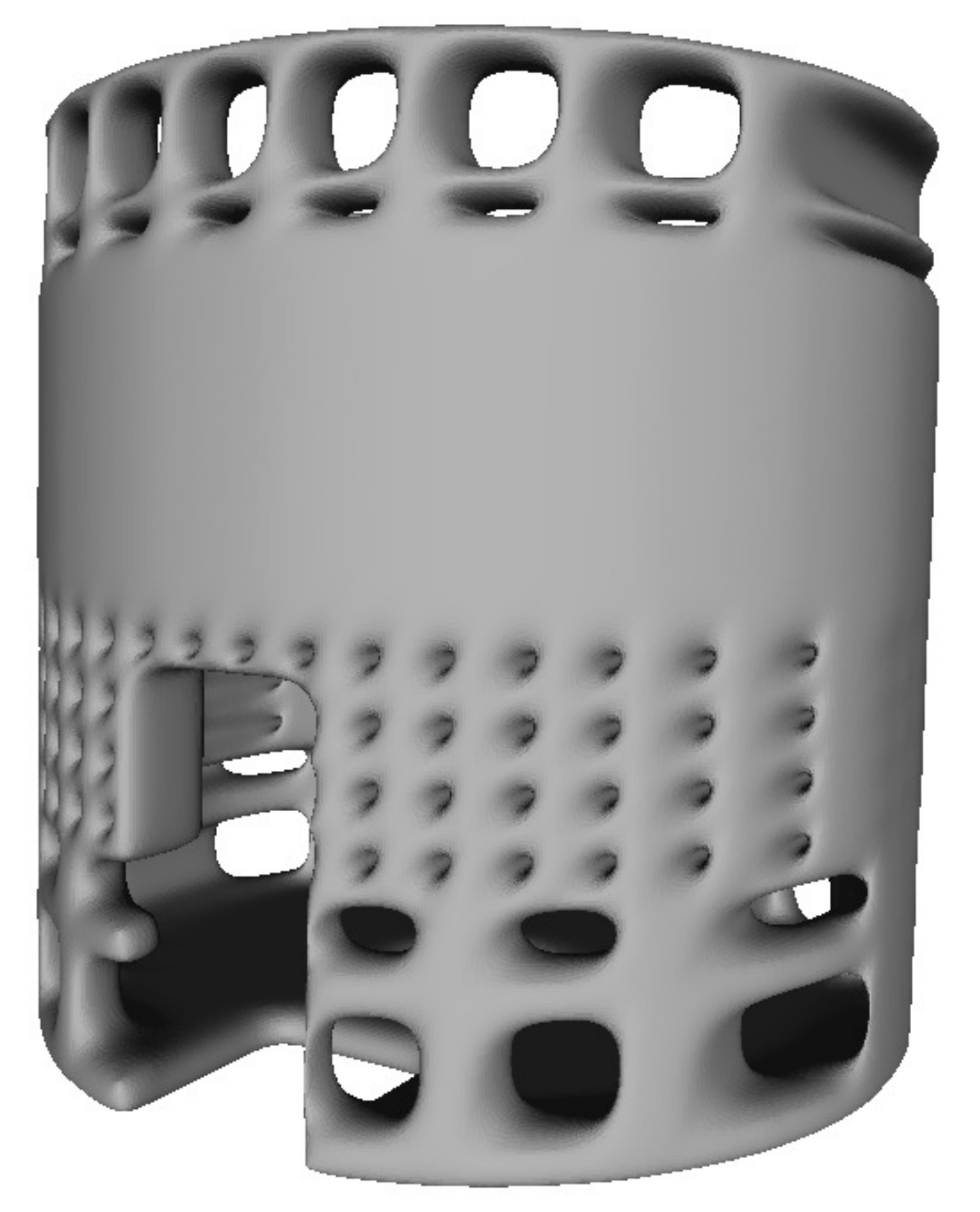}
	}
	\\
	\subfigure[Generation 9.] {
		\includegraphics[width=2.5cm]{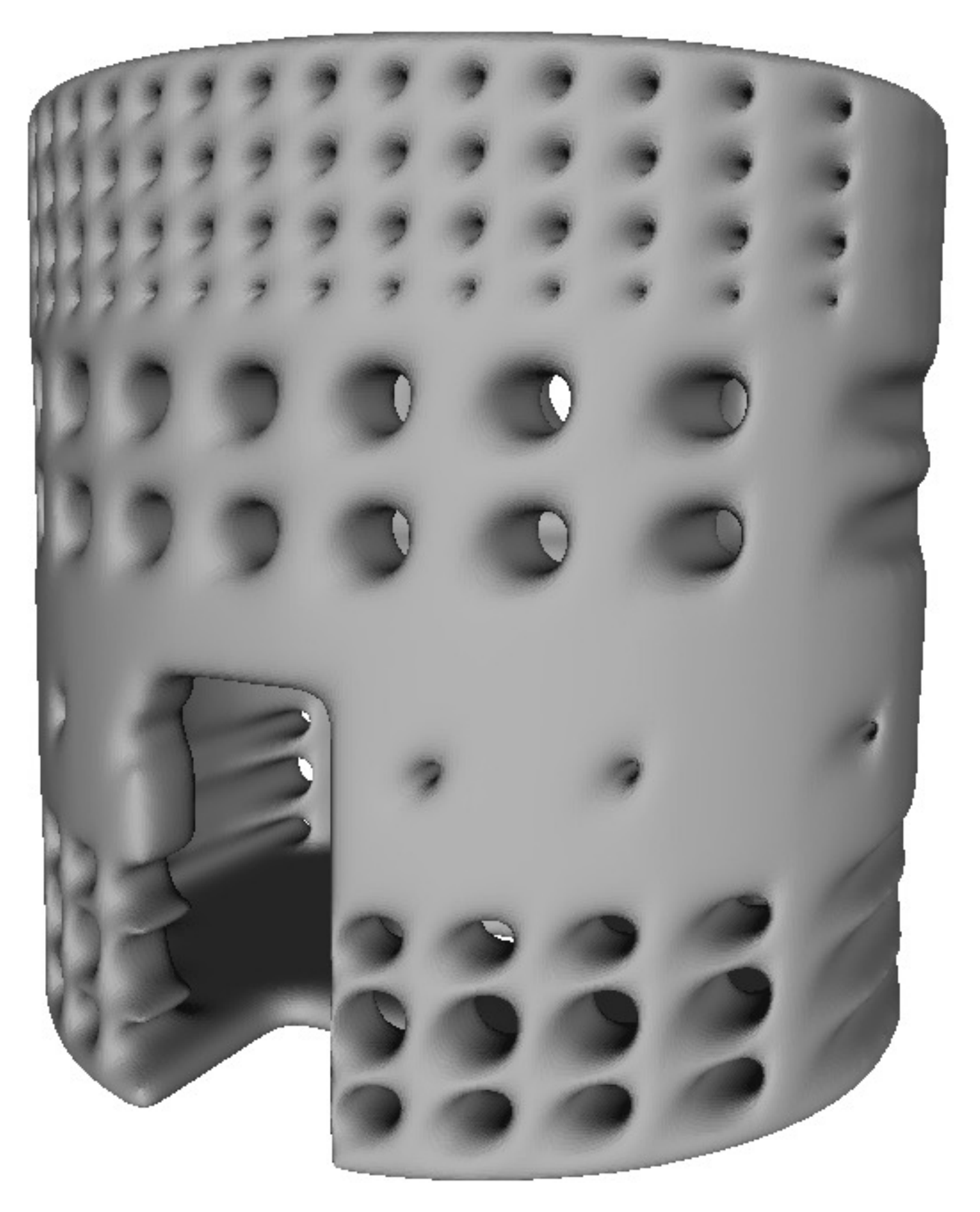}
		\includegraphics[width=2.5cm]{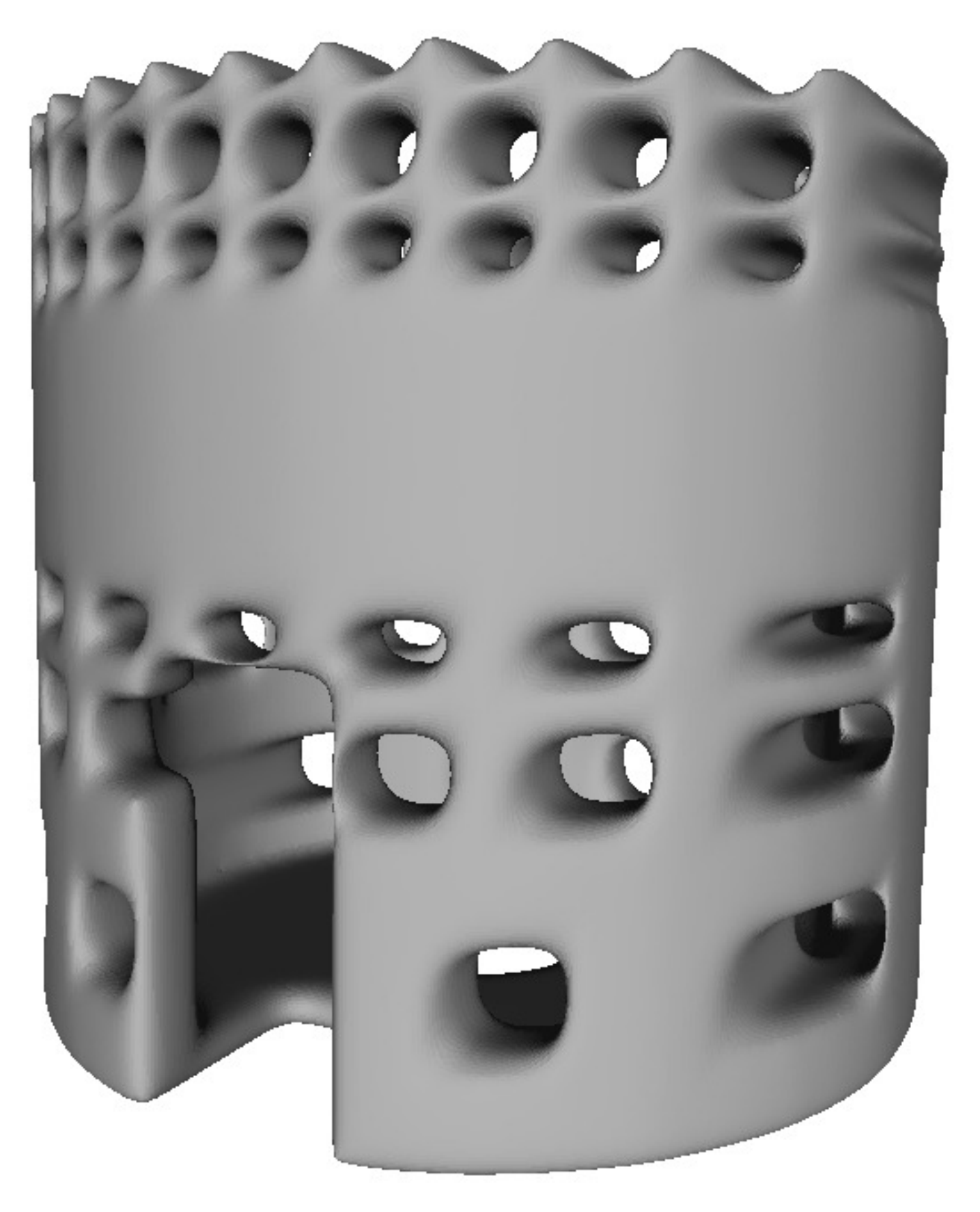}
		\includegraphics[width=2.5cm]{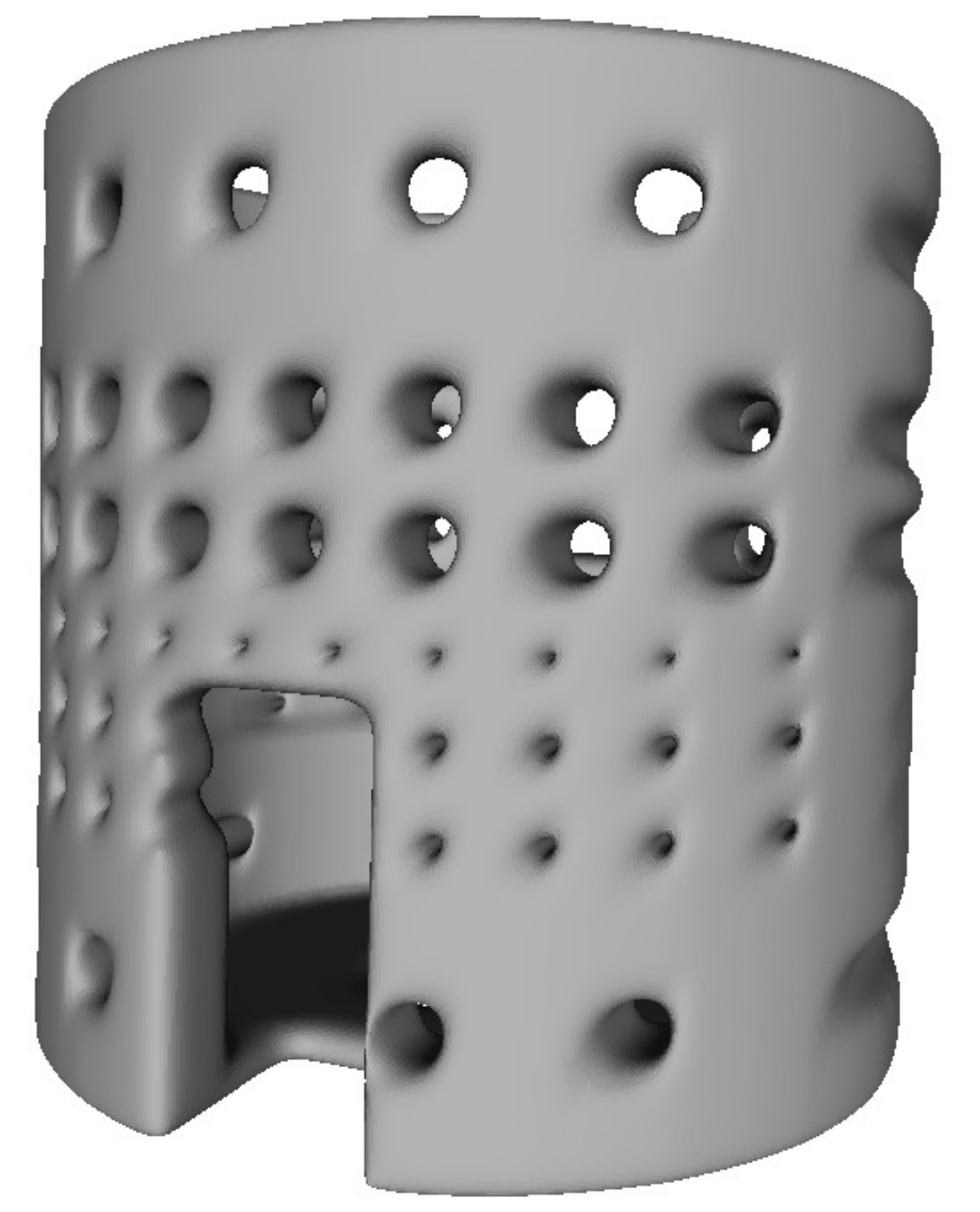}
		\includegraphics[width=2.5cm]{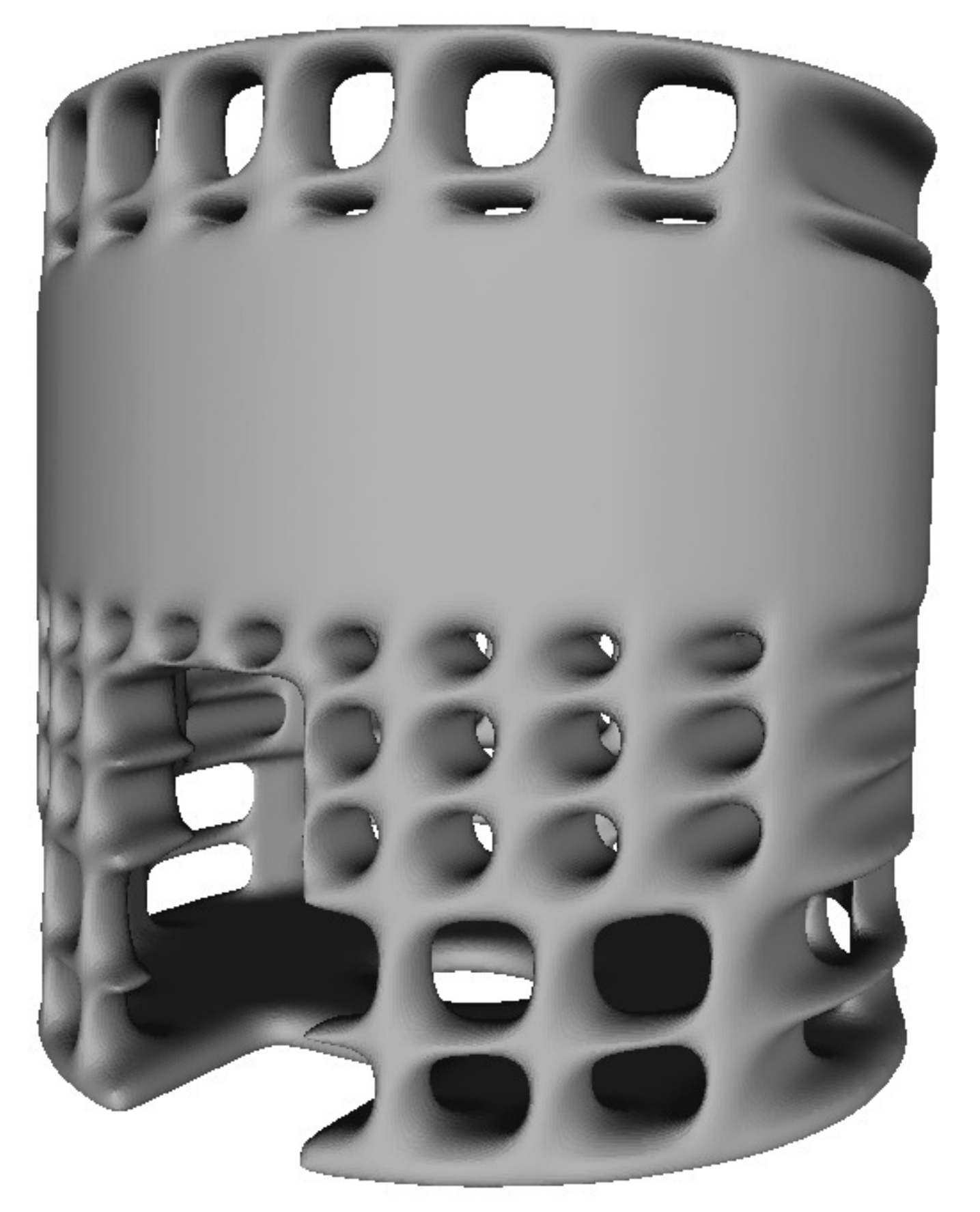}
	}
	\caption{Fittest evolved anodic chamber insert designs. From left to right, first position in the cascade to the last.}
	\label{fig:designs}
\end{figure}

After 10 generations the average power of an MFC unit in the fittest cascade had increased from 53.8 to 77.8~{\textmu}W (a 44.6\% increase) with an average MFC unit power density increase from 1.95 to 2.93~{\textmu}W/mL (a 50.3\% increase; volumetric power density was normalised by the anolyte volume.) The individual MFC unit power values from generation 0 and 9 can be seen in Fig.~\ref{fig:power} and the designs are shown in Fig.~\ref{fig:designs}. 

In comparison with the standard MFC cascade, i.e., without any 3-D printed inserts, the evolved cascade produced $\sim$20\% more power and double the power density. Treatment efficiency measured in COD values was over 90\% for both the fittest (93.2\%) and standard (94\%) MFC cascades. This is higher than the generation 0 cascades, which was 86.6\%. This demonstrates that system efficiency, in terms of both power generating performance and treatment efficiency, can be improved through design optimisation. When the inserts' MFC cascade position was inverted, i.e., the insert in the first position becoming the insert in the fourth, insert in the second position becoming the insert in the third, and vice-versa, a reduction in cascade power of 12.5\% was observed, showing that evolution has exploited characteristics specific to the cascade flow. In addition, the same insert designs from the tenth generation were fabricated using standard (non-conductive) PLA and a reduction in power of 10.4\% was seen, showing that the conductivity of the inserts had a positive effect on the MFC power generation by augmenting the carbon veil anode electrode. The results are summarised in Table~\ref{table:physical_res}.

\begin{table}[t]
	\caption{Physical MFC cascade coevolution results}
    \centering
	\begin{tabular}{L{2.50cm} L{2.50cm} L{2.50cm} L{2.50cm}}
		\toprule
		& Average Power Per Unit ({\textmu}W) & Average Power Density Per Unit ({\textmu}W/mL) & Stack Treatment Efficiency (\%) \\
		\toprule
		Gen 0 & 53.8 & 1.95 & 86.6 \\
		Gen 9 Inverted & 68.1 & 2.48 & 95.6 \\
		Gen 9 PLA & 69.7 & 2.49 & 88.6 \\
		Gen 9 & 77.8 & 2.93 & 93.2 \\
		No Inserts & 61.6 & 1.40 & 94.0 \\
		\bottomrule
    \end{tabular}
    \label{table:physical_res}
\end{table}
 
Overall, the total anolyte volume in the cascade remained unchanged from the first generation at 110~mL. However, the volume of anolyte was redistributed to the other MFC units from the third position where the insert increased in volume; i.e., all other inserts decreased in volume. The surface area of all insert designs increased and consequently the total stack surface area increased by 6.45~cm$^2$. A larger surface area is a desirable aspect for the MFC anode since it provides a larger space for electrochemically active bacteria attachment~\citep{You:2014}. Although the conductive PLA inserts used in this study were not used simply for expanding the size of anodes, it seems clear that the larger surface area of the conductive inserts contributed to the MFC power production. However, due to the relatively low conductivity of the filament material, it would be unlikely to replace the carbon fibre veil anode~\citep{You:2017}. 3-D printing structures that serve as both anodes and chassis will be an interesting avenue to explore in future work, in terms of additive layer manufacturing for scale-up.

The decrease in volume of the insert in the first position was a result of increases in the diameter of the holes in the bottom half of the design, resulting in an increase in anolyte volume from 25.8~mL to 26.6~mL. The total surface area of the insert increased by 1.2~cm$^2$ and its power production increased by 38.45~{\textmu}W. 

The MFC unit in the second position decreased in volume as a result of increases in the diameter of the holes and an increase to the inner radius of the middle sections, resulting in an increase in anolyte volume from 27.4~mL to 28.2~mL. The bottom and top sections of the insert remained unchanged after 10 generations. The total surface area of the second insert increased by 1.16~cm$^2$, however the amount of power produced remained unchanged.

The most significant changes were observed in the third position in the cascade where the offspring in this species produced a greater average MFC cascade power in 4 of the 10 evolved generations. The third insert increased in overall volume causing a reduction in the anolyte volume from 28.6~mL to 25.5~mL. The volume of the bottom 3 sections of the insert increased by 40\% through a smaller inner radius and a reduction in the number of holes, whereas the volume of the top quarter section decreased by 18\% through a larger inner radius causing a larger anolyte volume at the top near the unit outlet. The third section from the bottom of the first and third inserts experienced identical changes, shrinking the inner radius to its minimum (that is, filling as far inwards towards the carbon veil anode as permitted) and with holes of 2~mm diameter separated by 2~mm of filament resulting in a larger surface area. The total surface area of the third insert increased by 2.15~cm$^2$ and the power increased by 64~{\textmu}W.

\begin{figure}[t]
	\centering 
	\includegraphics[width=2.5cm]{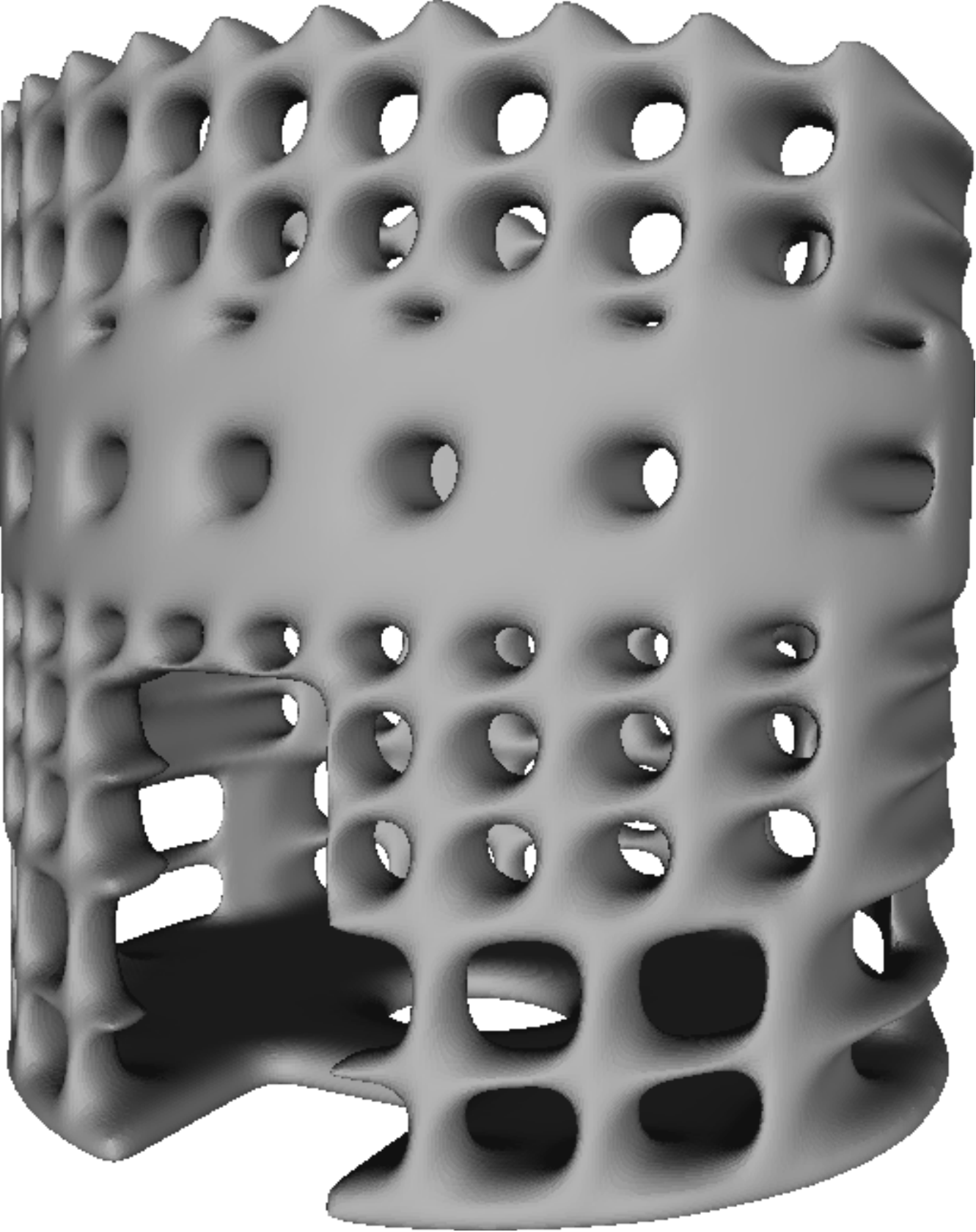}
	\caption{End species fittest evolved anodic chamber insert design.}
	\label{fig:end_design}
\end{figure}

Similar to the first position in the stack, the insert in the last position experienced a decrease in volume as a result of increases in the diameter of the holes in the bottom half of the design, with a resulting increase in anolyte volume from 28.2~mL to 29.9~mL. The total surface area increased by 1.94~cm$^2$. However, the power remained essentially unchanged, producing a small decrease of 6~{\textmu}W. Therefore, 5 additional generations were performed using Algorithm~\ref{alg:1plus4} solely on the last species, i.e., the first 3 MFC units were fixed to the fittest designs from generation 9. This resulted in a further increase of 18\% in the cascade power output. The final evolved insert design further reduced in volume as a consequence of additional holes to the upper half, resulting in a further increase in anolyte volume to 30.9~mL, showing a continued trend towards increasing the amount of feedstock in the end position; the design can be seen in Fig.~\ref{fig:end_design}.

\section{Conclusions}

The use of cascade configurations and 3-D printed components have recently been identified as significant ways through which MFCs can be enhanced~\citep{Walter:2016,Papaharalabos:2015}. The design and optimisation of the anode electrode has been highlighted as having a crucial affect on performance and scalability since it is central to biocatalysis through bacterial adhesion and electron transfer.

By repeatedly creating and testing physical designs as suggested by computational intelligence it may be possible to discover and exploit previously unknown or insufficiently understood physical interactions. In this paper, we have provided proof-of-concept that such an approach can be used to design conductive structures that augment the existing anodes in MFC units forming a cascade system. The structures were inserted into the anodic chamber facilitating the testing of new designs to explore the biocompatibility, conductivity, stability, surface area, and hydrodynamics. The total cascade performance in terms of power output and power density iteratively increased. Coevolution enables the designs in each unit to evolve towards the optimum characteristics specific to the cascade position whereby the total cascade performance is optimised.

Design mining provides a general and flexible approach to the optimisation of MFCs. Future work will therefore include the use of new 3-D printers and materials, such as those with a higher conductivity and the ability to control microsurfaces and create microporous objects. Multi-material printers will enable the production of monolithic MFCs enabling the optimisation of entire fuel cells within a cascade system.

\section*{Acknowledgement}

This work was supported by the Engineering and Physical Sciences Research Council under Grant EP/N005740/1.



\end{document}